\documentclass{article}

\usepackage{amssymb}
\usepackage{marvosym}
\usepackage{amsmath}
\usepackage{wrapfig}
\usepackage{graphicx}
\usepackage{enumitem}
\usepackage[preprint]{corl_2025} 

\usepackage{hyperref}
\usepackage{booktabs}
\usepackage{multirow}
\usepackage{colortbl}
\usepackage{rotating}

\usepackage{tabularx}
\usepackage{array}
\usepackage{ragged2e}

\usepackage[table]{xcolor}
\hypersetup{
    colorlinks=true,
    linkcolor=orange,
    filecolor=magenta,
    urlcolor=orange,
    citecolor=orange,
}

\title{RoboOS: A Hierarchical Embodied Framework for Cross-Embodiment and Multi-Agent Collaboration}

\author{
\textbf{Huajie Tan$^{1,2,*}$, Xiaoshuai Hao$^{2,*}$, Cheng Chi$^{2,*}$, Minglan Lin$^{2,*\dagger}$}, Yaoxu Lyu$^{1,2}$\\
\textbf{Mingyu Cao$^{2}$, Dong Liang$^{2}$, Zhuo Chen$^{2}$, Mengsi Lyu$^{2}$, Cheng Peng$^{2}$, Chenrui He$^{2}$} \\
\textbf{Yulong Ao$^2$, Yonghua Lin$^2$, Pengwei Wang$^{2,\dagger}$, Zhongyuan Wang$^2$, Shanghang Zhang$^{1,2,\text{\Letter}}$} \\
$^1$ State Key Laboratory of Multimedia Information Processing, \\
School of Computer Science, Peking University \\ 
$^2$ Beijing Academy of Artificial Intelligence (BAAI)
}

\begin{document}
\maketitle

\let\thefootnote\relax\footnotetext{$^{*}$ Equal contribution. $^{\dagger}$ Project leader. $^{\text{\Letter}}$ Corresponding author.}

\vspace{-1em}
\begin{abstract}
The dawn of embodied intelligence has ushered in an unprecedented imperative for resilient, cognition-enabled multi-agent collaboration across next-generation industrial ecosystems, revolutionizing paradigms in autonomous manufacturing, adaptive service robotics, and cyber-physical production architectures.
However, current robotic systems face significant limitations, such as limited cross-embodiment adaptability, inefficient task scheduling, and insufficient dynamic error correction.
While End-to-end Vision Language Action (VLA) models demonstrate inadequate long-horizon planning and task generalization, hierarchical VLA models suffer from a lack of cross-embodiment compatibility and multi-agent coordination capabilities.
To address these challenges, we introduce \textit{\textbf{RoboOS}}, the first open-source embodied system built on a \texttt{Brain-Cerebellum} hierarchical architecture, enabling a paradigm shift from single-agent to multi-agent intelligence.
Specifically, \textit{\textbf{RoboOS}} consists of three key components: \textit{\textbf{(1) the Embodied Brain Model (RoboBrain)}}, a multimodal large language model (MLLM) designed for global perception and high-level decision-making; \textit{\textbf{(2) the Cerebellum Skill Library}}, a modular, plug-and-play toolkit that facilitates seamless execution of multiple skills; and \textit{\textbf{(3) Real-Time Shared Memory}}, a spatiotemporal synchronization mechanism for coordinating multi-agent states.
By integrating hierarchical information flow, RoboOS bridges the Embodied Brain and the Cerebellum Skill Library, facilitating robust planning, scheduling, and error correction for long-horizon tasks, while ensuring efficient multi-agent collaboration through Real-Time Shared Memory.
Furthermore, we enhance edge-cloud communication and cloud-based distributed inference to facilitate high-frequency interactions and enable scalable deployment.
Extensive real-world experiments across various scenarios, such as \texttt{restaurants, households, and supermarket settings}, demonstrate RoboOS's versatility in supporting heterogeneous embodiments, including \texttt{single-arm, dual-arm, humanoid, and wheeled}.
This capability offers a scalable and practical solution for cross-embodiment collaboration, advancing the frontiers of embodied intelligence.
Project website: \href{https://github.com/FlagOpen/RoboOS}{RoboOS}.
\end{abstract}

\section{Introduction}
\label{sec:intro}

The rapid evolution of embodied intelligence has ushered in a transformative era for industrial automation, service robotics, and smart manufacturing, where robust multi-agent collaboration has become essential \cite{amazon_robotics_2025, mandi2024roco, an2023multi, liu2024coherent}.
Despite significant advancements, current robotic systems face persistent limitations, including poor cross-embodiment adaptability, inefficient task scheduling, and inadequate dynamic error correction.
While End-to-end Vision Language Action (VLA) models like OpenVLA~\cite{kim2024openvla}, RDT-1B~\cite{liu2024rdt}, and $\pi_{0}$~\cite{black2024pi_0} demonstrate weak long-horizon planning and task generalization, hierarchical VLA frameworks such as Helix~\cite{Helix2024}, Gemini-Robotics~\cite{team2025gemini_robo}, GR00T-N1~\cite{bjorck2025gr00t}, Hi-Robot~\cite{shi2025hi} and $\pi_{0.5}$~\cite{pi05} suffer from fragmented cross-embodiment compatibility and challenges in scalable multi-agent coordination. 
These issues highlight the urgent need for a unified system that bridges high-level cognition with low-latency execution while facilitating seamless collaboration among heterogeneous robots.

To address these gaps, we introduce \textit{\textbf{RoboOS}}, the first open-source embodied system built on a biologically inspired Brain-Cerebellum hierarchical architecture~\cite{israely2025cerebellar, ren2025neuronal, zhao2025dual}, representing a paradigm shift from single-agent to multi-agent intelligence. RoboOS innovatively incorporates three groundbreaking components: \textit{\textbf{(1) The Embodied Brain Model (RoboBrain~\cite{ji2025robobrain})}}, a multimodal large language model (MLLM) that orchestrates global perception—including 3D scene reconstruction and historical state tracking—and high-level decision-making for multi-agent task decomposition and affordance-aware trajectory generation, while dynamically correcting errors through real-time replanning; \textit{\textbf{(2) The Cerebellum Skill Library}}, a modular, plug-and-play toolkit that supports heterogeneous embodiments (\textit{e.g.}, single-arm, humanoid) with low-latency execution for manipulation (VLA-based tools~\cite{kim2024openvla,liu2024rdt,black2024pi_0}, 
Expert-based tools~\cite{tang2025affordgrasp,fang2023anygrasp}), navigation (VLN-based tools~\cite{zhang2025mapnav,zhang2024navid}, SLAM~\cite{xu2025fastliolocalization, grisetti2010tutorial}), and specialized skills ~\cite{fu2025cordvip,zhong2025dexgrasp,aksoy2025planning}; and \textit{\textbf{(3) Real-Time Shared Memory}}, a spatiotemporal synchronization hub that maintains spatial memory (\textit{e.g.}, spatial relationship, locations of objects and robots), temporal memory (\textit{e.g.}, task feedback, tool-calling history), and embodiment memory (\textit{e.g.}, motion domain, joint states and battery levels) to enable fault prediction and load balancing across robots.
\begin{wrapfigure}{r}{0.6\linewidth}
    \centering
    \vspace{-0.3cm}
    \includegraphics[width=\linewidth]{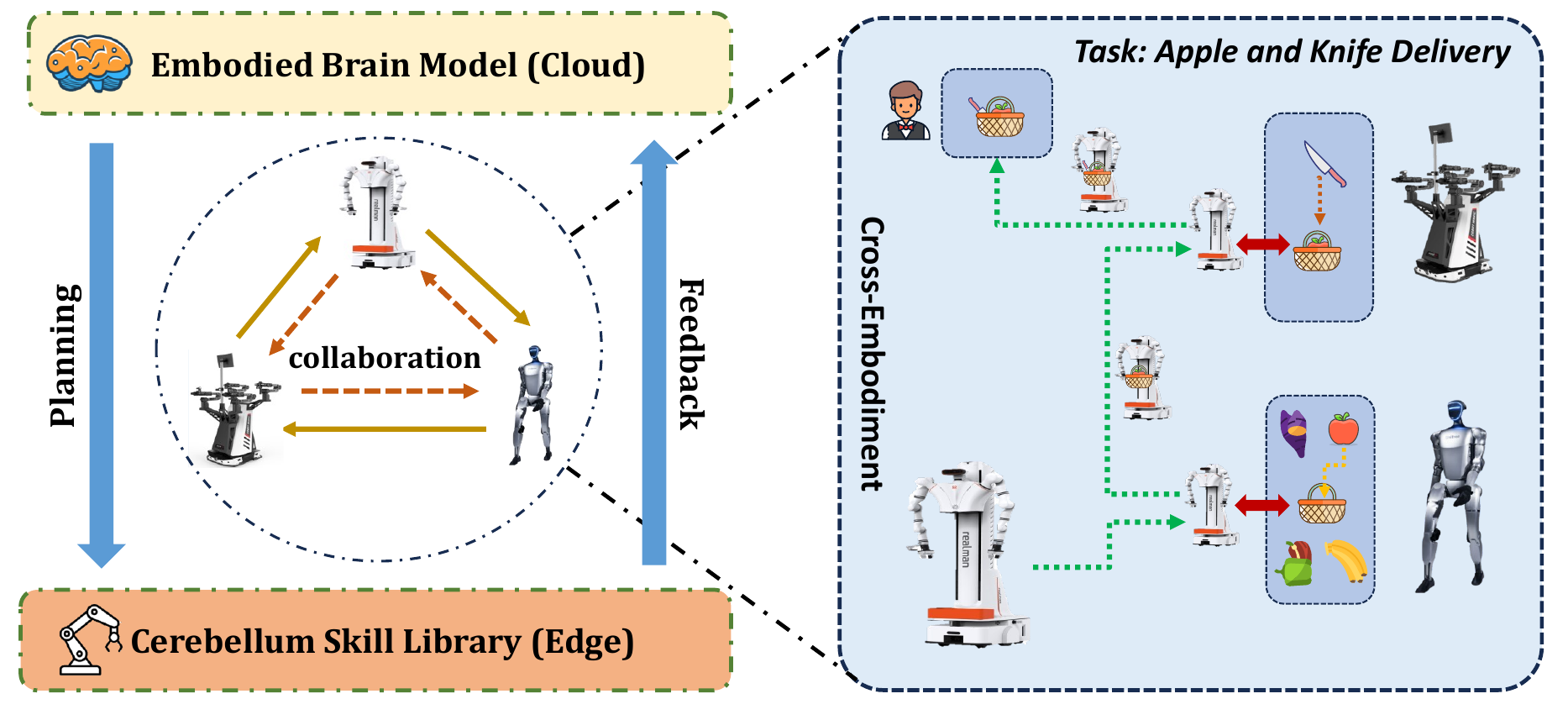}
    \caption{
        \footnotesize{\textbf{Overview of RoboOS.}}
    }
    \vspace{-0.4cm}
    \label{fig:teasor}
\end{wrapfigure}

Moreover, \textbf{\textit{RoboOS}} optimizes scalability through edge-cloud communication and cloud-based distributed inference, ensuring high-frequency interactions and large-scale deployment of cloud inference with our Flagscale framework~\cite{FlagScale2024}. 
Extensive real-world validation in diverse scenarios—from industrial assembly to household services—demonstrates its versatility across heterogeneous robots, including dual-arm manipulators and wheeled platforms.
For instance, to tackle a collaborative ``apple-and-knife delivery''  task, as shown in Fig.~\ref{fig:teasor}, RoboOS dynamically allocates subtasks to three distinct robots (\texttt{Unitree Humanoid, AgileX Dual-arm, and RealMan Single-arm}) via shared memory, achieving seamless coordination through RoboBrain’s task decomposition and the Cerebellum’s skill execution.

Our main contributions are summarized as follows:

\begin{itemize}[left=1em] 
\item
We propose \textbf{\textit{RoboOS}}, the pioneering open-source embodied system built on a \texttt{Brain-Cerebellum Hierarchical Architecture}, facilitating a transformative shift from single-agent systems to multi-agent intelligence.

\item 
We meticulously designed \textbf{\textit{RoboOS}} with three core components: the Embodied Brain Model, the Cerebellum Skill Library, and Real-Time Shared Memory. Additionally, we optimized it for efficient edge-cloud communication and distributed inference, enhancing its overall performance and scalability.
 
\item 
Extensive real-world experiments across various scenarios—including \texttt{restaurants, households, and supermarket settings}—validate RoboOS's adaptability and performance across diverse embodiments, such as \texttt{single-arm, dual-arm, humanoid, and wheeled robots}. This showcases its effectiveness in cross-embodiment collaboration and advances the capabilities of embodied intelligence, leading to practical and scalable solutions.
\end{itemize}

\clearpage
\section{Related Work}
\label{sec:related}
\textbf{Multimodal Large Language Models}  
Recent advancements in Vision-Language Models (VLMs) have demonstrated exceptional multimodal understanding. Proprietary models~\cite{hurst2024gpt4o, anthropic2024claude3.5, wuevaluating, Google2023gemini} and open-source alternatives~\cite{Qwen2.5-VL, chen2024internvl, li2024llavaov, abdin2024phi, beyer2024paligemma} have set new benchmarks in visual question answering (VQA), image captioning, and multimodal dialogue through large-scale pretraining on image-text pairs. 
Reasoning-enhanced models like GPT-o1~\cite{OpenAI2024o1}, DeepSeek-R1~\cite{guo2025deepseek}, and Kimi-1.5~\cite{team2025kimi} show that post-training reinforcement learning (RL) can significantly improve mathematical and coding abilities. RL-based reasoning MLLMs~\cite{tan2025reason, huang2025vision} also excel in multimodal reasoning tasks. 
However, transferring these capabilities to embodied intelligence systems is a challenge. Works like EmbodiedGPT~\cite{mu2023embodiedgpt} and RoboBrain~\cite{ji2025robobrain} explore integrating visual-language understanding with robot-specific skills, including long-horizon task planning and trajectory synthesis. Extending reasoning-enhanced MLLMs to embodied scenarios~\cite{tan2025reason,  zhang2025embodied} represents a promising research frontier.

\textbf{Vision-Language-Action (VLA) Models}  
Building on the capabilities of VLMs~\cite{ chen2024internvl, li2024llavaov, beyer2024paligemma, li2024foundation}, researchers have developed vision-language-action (VLA) models for robotic manipulation tasks. Current VLAs are categorized as follows:
End-to-End VLAs directly map visual-textual inputs to actions using unified architectures, employing regression-based~\cite{li2023vision, brohan2023rt, kim2024openvla, hao2025tla, liu2024robomamba}, diffusion-based~\cite{liu2024rdt, black2024pi_0, team2024octo}, and hybrid approaches~\cite{liu2025hybridvla}. While effective for short-term tasks, they struggle with long-horizon planning and generalization.
Hierarchical VLAs address these limitations by using model-level~\cite{liu2024self, Helix2024, team2025gemini_robo, bjorck2025gr00t} or task-level hierarchies~\cite{shi2025hi, pi05} to decompose long-horizon tasks into subtasks. However, they still face challenges in cross-embodiment compatibility and multi-agent coordination.
To tackle these issues, we introduce RoboOS, the first open-source hierarchical embodied framework that facilitates cross-embodiment and multi-agent collaboration, achieving structural decoupling while maintaining functional synergy between MLLMs and VLAs.

\textbf{Multi-Robot Collaboration}  
Multi-robot collaboration (MRC) has been extensively studied for various applications~\cite{amazon_robotics_2025, mandi2024roco, an2023multi, liu2024coherent,HAO2025103018}. Research focuses on coordination, communication, and task allocation to enhance efficiency~\cite{rizk2019cooperative, fierro2018multi}. MRC shows promise in automated warehousing~\cite{agrawal2023rtaw}, search and rescue~\cite{guo2023cross}, and environmental monitoring~\cite{edwards2023collaborative}. Learning methods like reinforcement and imitation learning further improve MRC~\cite{patino2023learning, liu2023guide, pereida2018data}.
Despite advancements, MRC faces challenges in cross-embodiment adaptation, task allocation, and dynamic planning, impacting real-time coordination. This paper introduces RoboOS, a hierarchical architecture that addresses these issues and enhances edge-cloud communication for scalable deployment.

\section{Method}
\label{sec:method}
In this section, we first introduce the framework of our proposed \textbf{\textit{RoboOS}} and explain the functionalities of its three key components. Next, we outline the primary workflow pipeline for multi-robot collaboration within \textbf{\textit{RoboOS}}, detailing the implementation of hierarchical information interaction. 
Finally, we present optimizations for edge-cloud communication and cloud-based distributed inference, ensuring high-frequency interactions and enabling large-scale deployment.

\begin{figure*}[t]
    \centering
   \vspace{-0.2cm}
    \includegraphics[width=0.97\linewidth]{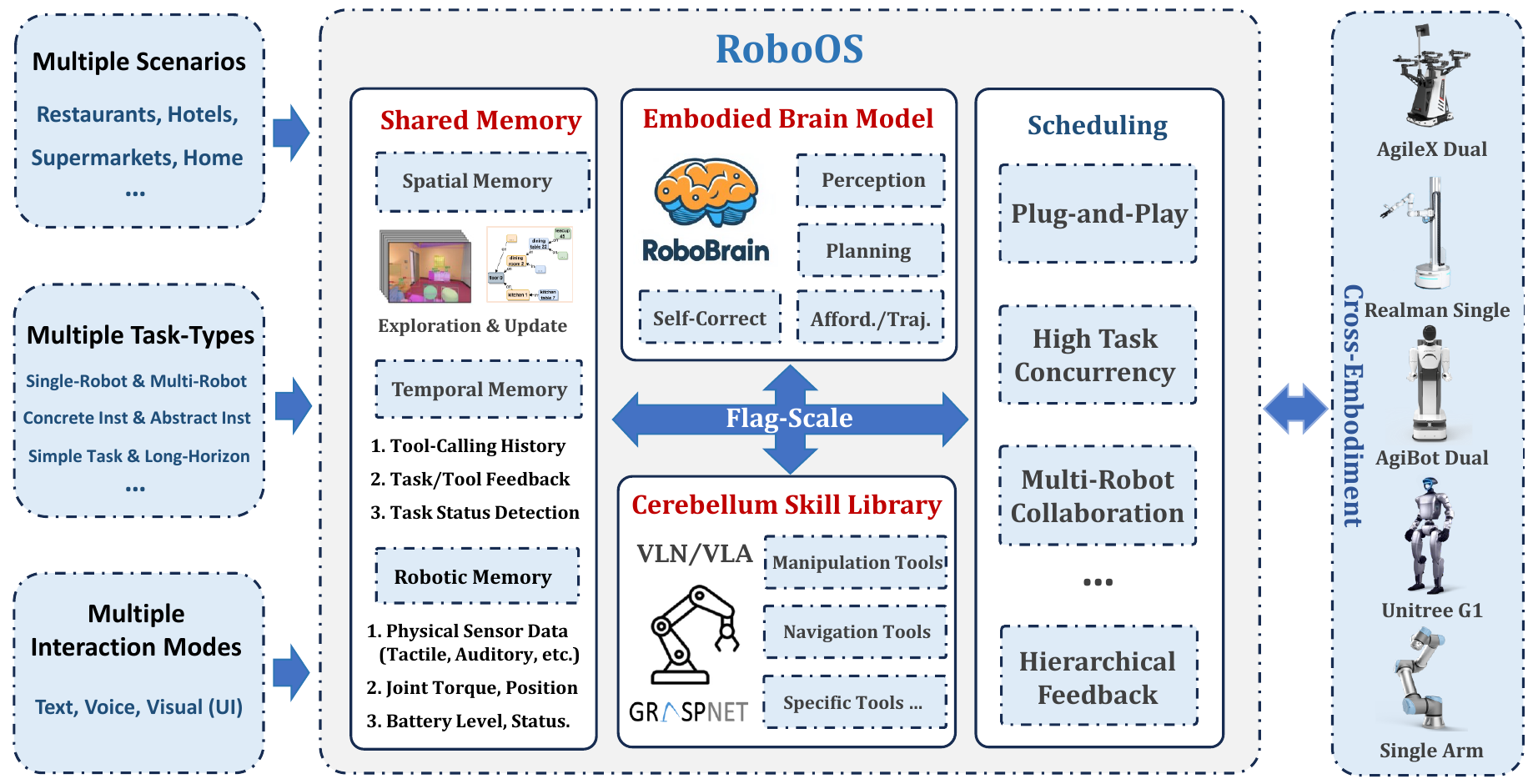}
    \caption{
    \textbf{Framework of RoboOS.} RoboOS is a Brain-Cerebellum hierarchical architecture for multi-robot collaboration that comprises three core components: a) Cloud-based Embodied Brain model for high-level cognition and multi-agent coordination; b) Distributed Cerebellum Modules for executing robot-specific skills; and c) Real-Time Shared Memory for enhanced environmental awareness.
        }
       \vspace{-0.5cm}
    \label{fig:method}
\end{figure*}

\subsection{Framework of RoboOS}
\label{subsec:framework}
As shown in Fig.~\ref{fig:method}, \textbf{\textit{RoboOS}} is a unified embodied system built on a biologically inspired Brain-Cerebellum hierarchical architecture, comprising three core components: the Embodied Brain Model (\texttt{RoboBrain~\cite{ji2025robobrain}}), the Cerebellum Skill Library, and Real-Time Shared Memory. Deployed via the FlagScale MLLMs toolkit~\cite{FlagScale2024}, the edge-cloud RoboOS framework facilitates seamless multi-robot coordination by synchronizing cognition across the multiple agents.
The system operates as follows: First, the Embodied Brain Model manages system-wide tasks, including multi-robot task planning, tool invocation, spatiotemporal memory updates, and adaptive error correction through continuous three-level feedback loops. Second, the Cerebellum Skill Library, deployed on individual robot terminals, offers modular, plug-and-play functionality via standardized Robot Profiles. Finally, the Redis-optimized Shared Memory maintains a dynamic knowledge base of spatial relationships, operational states, and historical data to support real-time decision making. This architecture ensures robust large-scale deployment while maintaining the low-latency interactions essential for embodied AI systems.

\textbf{Embodied Brain Model (RoboBrain)} 
The cloud-deployed MLLM can be any existing model, including proprietary models~\cite{hurst2024gpt4o, anthropic2024claude3.5, Google2023gemini} or open-source options~\cite{liu2024deepseek,Qwen2.5-VL,chen2024internvl,li2024llavaov,abdin2024phi,beyer2024paligemma}. To better suit embodied scenarios, we adopt RoboBrain~\cite{ji2025robobrain} as our Embodied Brain Model, enhancing its capabilities for the RoboOS framework. Building on RoboBrain's functionalities—single-robot planning, affordance prediction, and trajectory prediction—we implement multi-stage training to improve multi-robot task planning, agent-based tool invocation, and spatiotemporal updates using the pretrained Qwen2.5VL-7B model~\cite{Qwen2.5-VL}. Key enhancements include \textit{Multi-Robot Task Planning}, utilizing real-time shared spatiotemporal memory to predict workflow topologies for collaborative tasks; \textit{Agent-Based Tool Invocation}, managing agents and invoking tools as needed with self-corrective planning based on feedback; \textit{Spatiotemporal Memory Update}, dynamically updating shared memory in real time according to subtask execution and tool feedback; and \textit{Low-Level Guidance}, predicting manipulable regions and trajectories during tool execution to assist in manipulation~\cite{ji2025robobrain}.

\textbf{Cerebellum Skill Library} 
This modular, plug-and-play embodied toolkit supports various robotic embodiments (\textit{e.g.}, single-arm, dual-arm, wheeled, humanoid) with low-latency execution for manipulation and navigation throughout the robotic task cycle. 
The Cerebellum Skill Library encompasses three key aspects: manipulation types, integrating both expert-based tools (\textit{e.g.}, affordance-aware grasping~\cite{tang2025affordgrasp}, generalized grasping~\cite{fang2023anygrasp}) and VLA-based tools (\textit{e.g.}, OpenVLA~\cite{kim2024openvla}, RDT-1B~\cite{liu2024rdt}, Octo~\cite{team2024octo}, $\pi_0$~\cite{black2024pi_0}); 
navigation types, supporting traditional mapping-localization-navigation pipelines (\textit{e.g.}, SLAM~\cite{xu2025fastliolocalization, grisetti2010tutorial}) and vision-language-navigation (VLN) tools (\textit{e.g.}, MapNav~\cite{zhang2025mapnav}, NavID~\cite{zhang2024navid}); 
and specialized skills for contact-rich interactions, deformable object handling, and dexterous hand control~\cite{fu2025cordvip, zhong2025dexgrasp, aksoy2025planning}. 
Standardized tool and robot profiles ensure seamless integration and interoperability across diverse robotic platforms.

\textbf{Real-Time Shared Memory} 
This component maintains spatial, temporal, and robotic memory to enable robust coordination and adaptive decision-making.  
\textit{Spatial memory} is represented as a dynamic scene graph~\cite{chang2021comprehensive}, structured hierarchically into floors, rooms, and object-level nodes~\cite{werby2024hovsg,gu2024conceptgraphs}. Each node encodes attributes such as position, affordances, and semantic labels, while edges represent spatial or functional relationships. Scene graphs are constructed from multi-view RGB-D inputs using SAM segmentation and CLIP features, merged via geometric clustering and open-vocabulary grounding.  
\textit{Temporal memory} records task execution history, feedback, tool-calling logs, and other temporal data to support adaptive decision-making.
\textit{Robotic memory} stores real-time system attributes such as motion domain constraints, joint states, and battery levels, optimizing task allocation based on each robot's capabilities, power status, and connectivity conditions.

\begin{figure*}[t]
    \centering
    \vspace{-0.2cm}
    \includegraphics[width=0.97\linewidth]{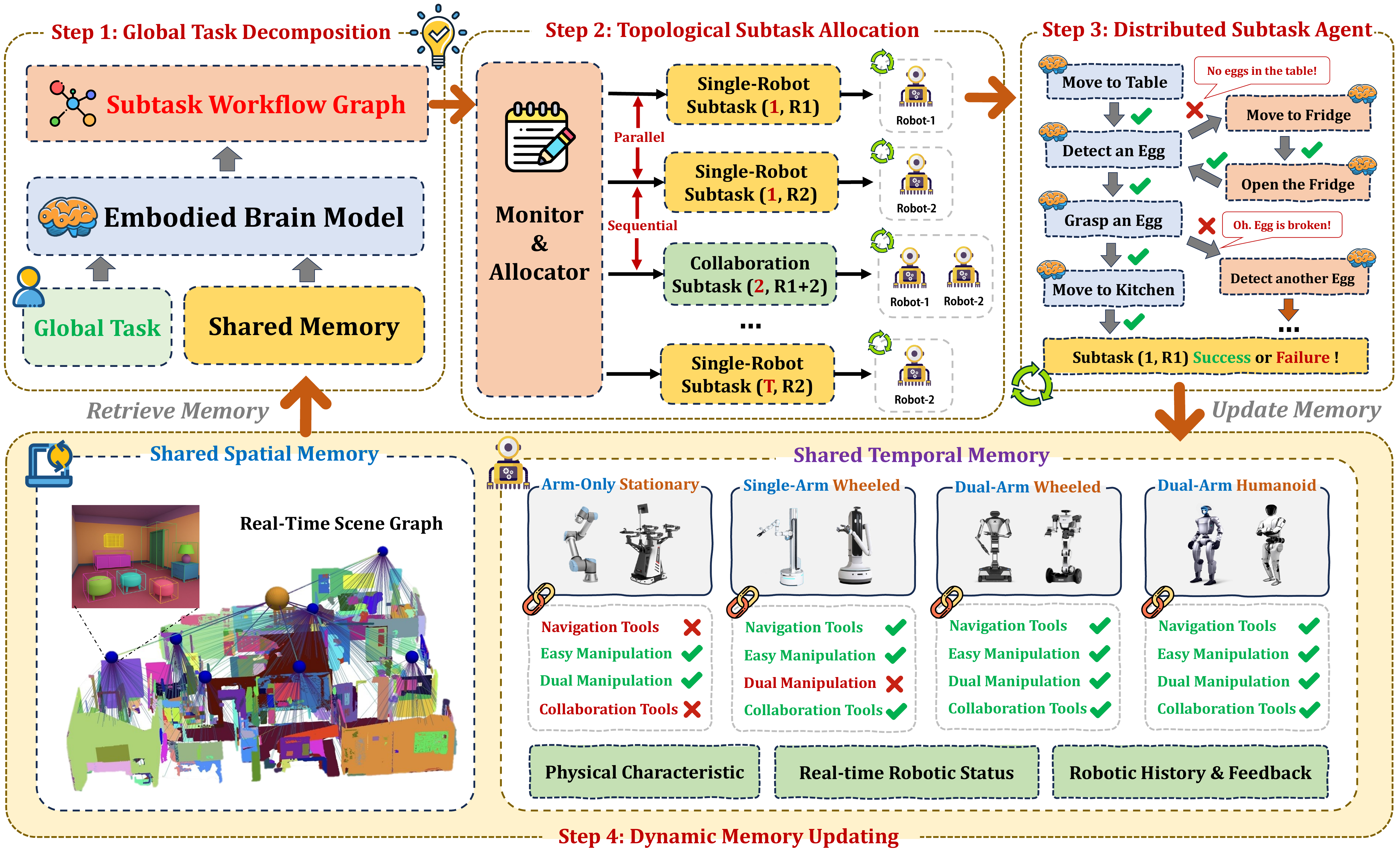}
    \caption{
    \textbf{Pipeline of RoboOS.} The RoboOS framework implements a workflow pipeline for multi-robot collaboration, consisting of four key phases: (1) hierarchical task decomposition, (2) topology-aware subtask allocation, (3) distributed agent-based execution, and (4) dynamic memory updating. This integrated workflow enables coordinated task completion while maintaining adaptability to environmental and operational constraints.
    }
       \vspace{-0.2cm}
    \label{fig:pipeline}
\end{figure*}

\subsection{Workflow Pipeline of RoboOS}
\label{subsec:pipeline}
The proposed RoboOS demonstrates high task concurrency and flexibility in multi-robot task allocation. To clarify the overall workflow pipeline of RoboOS, we use a single global task for detailed elaboration, as shown in Fig. \ref{fig:pipeline}.

\textbf{Step 1: Global Task Decomposition} Upon receiving the global task instruction \( T_{\text{global}} \), RoboOS initiates a Retrieval-Augmented Generation (RAG) process via RoboBrain to query the shared spatial memory, extracting environment-relevant information \( M_s \). This is integrated with (i) state feedback \( M_t \) from prior task executions (stored in shared temporal memory), (ii) the robot’s operational status \( S_r \) (\texttt{idle}, \texttt{busy}, or \texttt{offline}), (iii) the robot skill library \( M_r \), and (iv) \( T_{\text{global}} \). RoboBrain processes these inputs to generate a structured reasoning trace \( \mathcal{R} \) and a subtask graph \( \mathcal{G} \), formalized as: 
\begin{equation}
    (\mathcal{R}, \mathcal{G}) = \text{RoboBrain}\big(M_s \oplus M_t \oplus S_r \oplus M_r \oplus T_{\text{global}}\big),
\end{equation}
where \( \oplus \) denotes the concatenation or fusion of multimodal inputs.

\textbf{Step 2: Topological Subtask Allocation} 
The Monitor dynamically schedules and allocates subtasks in parallel based on the topological dependencies encoded in the directed acyclic graph \( \mathcal{G} \). 
Each subtask in \( \mathcal{G} \) is classified into two types: \textit{(1) Single-Robot Subtask \((d, r_i)\)}, executed autonomously by robot \( r_i \) at topological depth \( d \); and \textit{(2) Collaboration Subtask \((d, r_{i:j})\)}, requiring coordinated execution among multiple robots \(\{r_i, \dots, r_j\}\) at depth \( d \). Here, \( d \) represents the execution priority, while \( r_i \) (or \( r_{i:j} \)) denotes the assigned robot(s). 
To enforce dependency constraints, the Monitor employs \textit{\textbf{Parallel Allocation}}—executing independent subtasks concurrently at the same depth (\textit{e.g.}, \((1, R_1)\) and \((1, R_2)\) in Fig. \ref{fig:pipeline})—and \textit{\textbf{Sequential Allocation}}, where subtask \((d+1, r_k)\) is blocked until all prerequisites at depth \( d \) are fulfilled (\textit{e.g.}, \((2, R_{1+2})\)). 
In practice, the system supports concurrent management of multiple subtask graphs \(\{ \mathcal{G}_1, \mathcal{G}_2, \dots, \mathcal{G}_n \}\), ensuring real-time adaptability to dynamic robot states and evolving task dependencies.

\textbf{Step 3: Distributed Subtask Agent}  
For each subtask, RoboOS deploys a dedicated \textit{\textbf{Robotic Agent}} to manage execution. The Agent autonomously orchestrates tool selection from the Skill Library based on: (1) feedback from prior executions, (2) tool-calling history, and (3) partial spatial memory of the environment. This closed-loop reasoning facilitates dynamic error recovery.  
For example (Fig. \ref{fig:pipeline}), when tasked with ``Search for an egg and place it on the table'', the Agent sequentially invokes tools (\textit{e.g., ``detect an egg''}). If the search fails (\textit{e.g., no egg detected in the kitchen}), the Agent uses spatial memory to infer potential locations (\textit{e.g., the fridge}) and selects the navigation tool to ``move to fridge'', showcasing adaptive recovery through iterative tool refinement.

\textbf{Step 4: Dynamic Memory Updating}  
Scene graphs are updated incrementally as robots perceive and act. New observations or interaction effects (e.g., filling, moving) modify node attributes and spatial edges in real time. For instance, after placing a cup on a shelf, its position and supporting surface edge are updated. Occlusion and history-aware correction ensure continuity, even when objects are temporarily hidden.  
Additionally, feedback, tool-calling history, and robot states are logged in the Temporal Memory and Robotic Memory.

\subsection{Edge-Cloud Deployment}
\label{subsec:edge-cloud}
Built upon our parallel training and inference framework, FlagScale \cite{FlagScale2024}, RoboOS supports end-cloud collaboration for multi-robot systems, creating a unified foundation for embodied intelligence. Designed for ``multi-robot, multi-modal, multi-task'' scenarios, it offers exceptional scalability and ultra-low-latency responsiveness. In edge deployments, robots automatically establish bidirectional communication with the cloud-based RoboBrain upon registration, enabling real-time task scheduling and status feedback via an efficient publish-subscribe mechanism (average command response latency $< 0.001 \, \text{s}$).
To manage the vast perception and behavioral data generated during long-term operations, FlagScale includes a memory-optimized data access engine that supports TB-level historical data with in-memory random access. This facilitates task replay, anomaly backtracking, cross-task knowledge transfer, and other critical scenarios. Additionally, FlagScale's framework supports parallel inference and multi-task cooperative scheduling of large models across distributed devices, unlocking the systemic potential of RoboBrain.
\section{Experiment}
\label{sec:exp}

\subsection{Implementation Details}
\begin{wrapfigure}{r}{0.44\linewidth}
    \centering
    \vspace{-8.3em}
    \includegraphics[width=\linewidth]{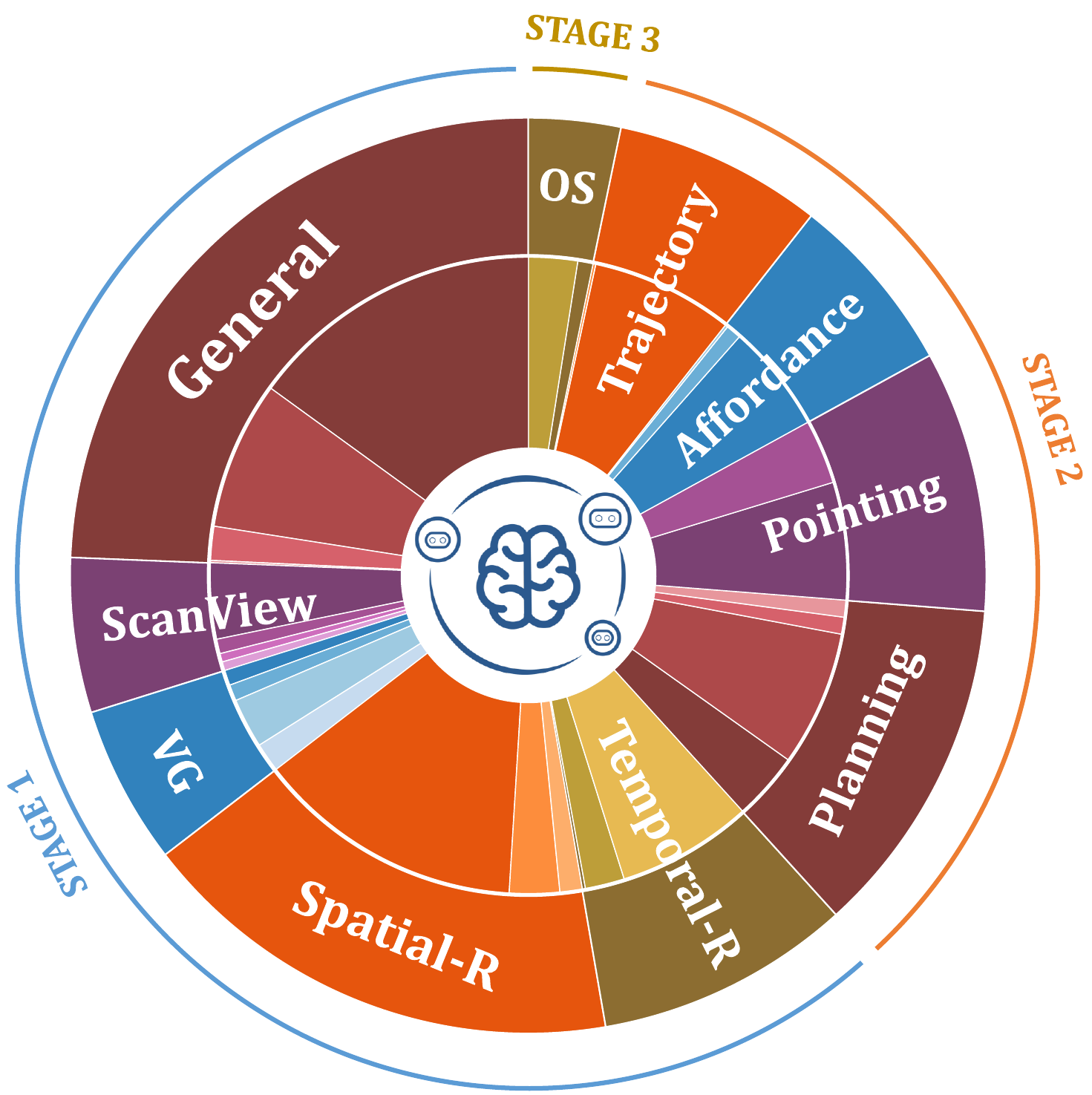}
    \caption{
        \footnotesize{\textbf{Training dataset distribution.}}
    }
    \vspace{-0.3cm}
    \label{fig:data}
\end{wrapfigure}
\textbf{Dataset Details} 
As shown in Fig.~\ref{fig:data}, the dataset for training the RoboBrain-1.5-OS model from pretrained Qwen2.5-VL-7B \cite{Qwen2.5-VL} consists of three categories: VLM datasets, Robotic datasets, and RoboOS-Enhanced datasets.
\textit{\textbf{(1) VLM Datasets:}} These are organized by capability type: General-873k \cite{li2024llavaov,liu2023mitigating} for enhancing general QA capabilities; ScanView-318k \cite{mmscan,leo,3rscan,scanqa,sqa3d} for improving multi-perspective scene perception; VG-326k \cite{li2024llavaov,yu2016modeling,mao2016generation,chen2024revisiting,krishna2017visual} for boosting visual grounding in object localization; Spatial-R-1005k \cite{yang2020graph,chen2020cops,tan2025reason} for spatial reasoning; and Temporal-R-525k \cite{zhang2025embodied,wan2022handmethat} for temporal reasoning. All data underwent rigorous cleaning to ensure the model retains strong QA abilities while enhancing localization and spatiotemporal reasoning.
\textit{\textbf{(2) Robotic Datasets:}} These were curated to target four core robotic operation capabilities: Planning, Pointing, Affordance, and Trajectory. Specifically, Planning-700k \cite{sermanet2024robovqa,ji2025robobrain,bu2025agibot} enhances long-horizon task planning; Pointing-537k \cite{deitke2024molmo,yuan2024robopoint} improves spatial position perception; Affordance-373k \cite{yuan2024robopoint,ramanathan2023paco,ji2025robobrain} predicts interactive object affordance regions; and Trajectory-428k \cite{niu2024llarva,ji2025robobrain} anticipates complete manipulation trajectories for successful execution.
\textit{\textbf{(3) RoboOS-Enhanced (OS) Datasets:}} We \textit{Multi-Robot Task Planning} and \textit{Agent-Based Tool Invocation} within the RoboOS framework. Specifically, we designed 68 multi-robot collaboration task types across supermarket, household, and restaurant scenarios, generating 45,000 samples using DeepSeek-V3 \cite{liu2024deepseek}. This dataset, named Multi-Robot-45k, features instances where each question includes a detailed scene graph, robot specifications, and a long-horizon collaborative task, while the corresponding answers provide reasoning processes and workflow graphs of decomposed subtasks. Additionally, we constructed Robotic-Agent-144k by generating correct Observation-Action pairs (positive samples) alongside probabilistically sampled error-injected Observation-Action pairs (negative samples) for each subtask from Multi-Robot-45k.

\textbf{Training Strategy}  
The training of the RoboBrain-1.5-OS model consists of three stages, as illustrated in Fig.~\ref{fig:data}. In STAGE-1, we utilize large-scale, high-quality VLM datasets with 3M samples to enhance foundational perception and reasoning. STAGE-2 employs carefully sampled robotic datasets to improve the model's four core embodied capabilities, incorporating 10\% of STAGE-1 data to prevent catastrophic forgetting, resulting in 2.3M samples. Finally, in STAGE-3, we apply RoboOS-Enhanced datasets for adaptability, mixing 2\% of STAGE-1 and 3\% of STAGE-2 data, yielding 249k samples. Throughout training, we used the Zero3 \cite{zero} distributed strategy on 20 servers, each equipped with 8$\times$A800 GPUs, with further details available in the appendix.

\textbf{Evaluation Metrics} For multi-robot planning, we use Accuracy-Rate (AR)~\cite{zhang2024lmms} to evaluate agent-based tool-calling in RoboOS. For pointing prediction, we employ the Where2Place benchmark~\cite{yuan2024robopoint} with AR to measure hit accuracy against target masks. For affordance and trajectory prediction, we evaluate using ShareRobot~\cite{ji2025robobrain} benchmark, where affordance is measured by mAP across IoU thresholds, while trajectory prediction uses Discrete Fréchet Distance (DFD)~\cite{gu2023rt}, Hausdorff Distance (HD), and RMSE for macroscopic and microscopic analysis. To better evaluate autonomous trajectory prediction, we removed the initial start-point hints during trajectory assessment.

\subsection{Results on Embodied Evaluation}
To evaluate the embodied capabilities of RoboBrain-1.5-OS—the core component of RoboOS—we selected comparable VLMs (\textit{e.g.}, LLaVA-OneVision-7B~\cite{li2024llavaov}, Qwen2.5-VL-7B~\cite{Qwen2.5-VL}) with similar parameter scales, alongside larger LLMs (\textit{e.g.}, Qwen3-14B~\cite{qwen3_2025}, DeepSeek-V3-685B~\cite{liu2024deepseek}) as general baselines. We also compared against embodied baselines such as RoboPoint-14B~\cite{yuan2024robopoint} and RoboBrain-1.0~\cite{ji2025robobrain}. As shown in Tab.~\ref{tab:main_tab}, RoboBrain-1.5-OS achieves outstanding performance in multi-robot planning, surpassing Qwen2.5-VL-7B by 28.14\% and outpacing DeepSeek-V3-685B by 5.53\%, thereby enhancing RoboOS's capabilities. It also outperforms all baselines in pointing, affordance, and trajectory prediction, improving over RoboBrain-1.0 by 3.64\%, 16.96\%, and 40.77\%, respectively, demonstrating superior results across multiple embodied capabilities.

\begin{table*}[t]
    \centering
    \caption{\textbf{Performance across four key embodied capabilities.} Top results are highlighted in \textbf{bold}.}
    \vspace{0.2cm}
    \resizebox{\textwidth}{!}{
    \begin{tabular}{l|cccc|ccc|c|ccc}
        \toprule
        \multicolumn{1}{l|}{\multirow{2}{*}{\textbf{Models / Metrics}}} & \multicolumn{4}{c|}{\textbf{Multi-Robot Planning}} & \multicolumn{3}{c|}{\textbf{Pointing}} &  \multicolumn{1}{c|}{\textbf{Affordance}} &  \multicolumn{3}{c}{\textbf{Trajectory}}\\ 
        \cmidrule(lr){2-5} \cmidrule(lr){6-8} \cmidrule(lr){9-9} \cmidrule(lr){10-12}
        & \textbf{Rest.} & \textbf{House.} &\textbf{Super.} &\textbf{AVG $\uparrow$} & \textbf{Seen} & \textbf{Unseen} & \textbf{AVG $\uparrow$} & \textbf{mAP $\uparrow$} & \textbf{DFD $\downarrow$} & \textbf{HD $\downarrow$} & \textbf{RMSE $\downarrow$} \\ 
        \midrule
        \rowcolor[HTML]{F2F2F2} \multicolumn{12}{l}{\textbf{General Baselines}} \\ \midrule
        Llava-OneVision-7B~\cite{li2024llavaov}             & 11.31 & 8.26 & 9.33 & 9.63 & 55.54 & 48.48 & 53.42 & 11.37 & 0.3558 & 0.3310 & 0.2749 \\
        Qwen2.5-VL-7B~\cite{Qwen2.5-VL}           & 43.22 & 59.30 & 58.29 & 53.60 & 57.20 & 47.60 & 54.32 & 14.06 & 0.2964 & 0.2751 & 0.2254 \\
        Qwen3-14B~\cite{qwen3_2025} & 47.74  & 63.82  & 43.22  & 51.60  & $\times$ & $\times$ & $\times$ & $\times$ & $\times$ & $\times$  & $\times$ \\
        DeepSeek-V3-685B~\cite{liu2024deepseek} & 69.85 & 83.92 & 74.87 & 76.21  & $\times$ & $\times$ & $\times$ & $\times$ & $\times$ & $\times$  & $\times$ \\ \midrule
        \rowcolor[HTML]{F2F2F2} \multicolumn{12}{l}{\textbf{Embodied Baselines}} \\ \midrule
        RoboPoint-14B~\cite{yuan2024robopoint} & -- & -- & -- & -- & 46.77 & 44.48 & 46.08 & -- & --  & -- & -- \\ 
        RoboBrain-7B-1.0~\cite{ji2025robobrain} & 17.59 & 12.06 & 10.55 & 13.40 & 54.64 & 49.45 & 53.09  & 27.10 & 0.1910 & 0.1710 & 0.1330 \\ 
        \rowcolor[HTML]{DAEFF9} RoboBrain-7B-1.5-OS & \textbf{78.39} & \textbf{86.93} & \textbf{79.90} & \textbf{81.74} & \textbf{57.23} & \textbf{55.57} & \textbf{56.73} & \textbf{44.06} & \textbf{0.0994} & \textbf{0.0966} & \textbf{0.0801} \\
        \bottomrule
    \end{tabular}
    }
    \label{tab:main_tab}
\vspace{-0.5em}
\end{table*}

\subsection{Demos on Real-World Collaboration}  
\begin{figure*}[t]
    \centering
    \vspace{-0.2cm}
    \includegraphics[width=0.98\linewidth]{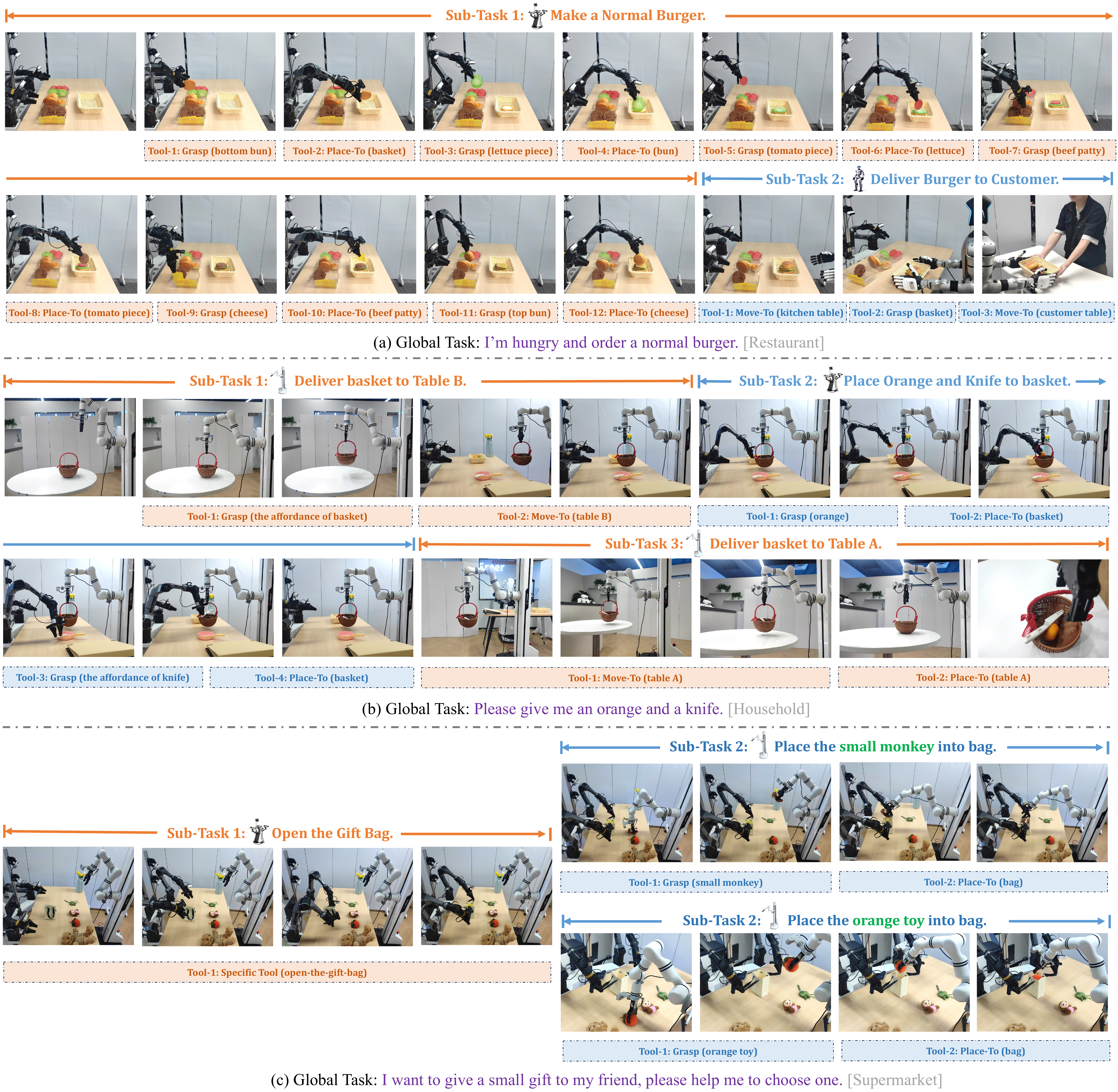}
    \vspace{-0.1cm}
    \caption{
\textbf{Real-world RoboOS Demonstrations.} We showcase multi-robot collaboration in three scenarios: (a) Restaurant: Unitree G1 and Agilex robots prepare burgers. (b) Household: Realman and Agilex robots fetch items. (c) Supermarket: Robots coordinate gift selection and packaging.
    }
       \vspace{-0.5cm}
    \label{fig:demo}
\end{figure*}

To demonstrate RoboOS's multi-robot collaboration, we present demos in restaurant, household, and supermarket settings. In the restaurant scenario (Fig.~\ref{fig:demo}~(a)), a Unitree G1 humanoid robot and an Agilex dual-arm robot work together to fulfill the task: ``I'm hungry and order a normal burger.'' RoboBrain-1.5-OS handles scene-aware reasoning, decomposing the task into subtasks for burger preparation and delivery.
In the household scenario (Fig.~\ref{fig:demo}~(b)), a Realman single-arm robot and an Agilex dual-arm robot collaborate to accomplish tasks like ``Give me an orange and a knife.'' 
In the supermarket (Fig.~\ref{fig:demo}~(c)), RoboBrain-1.5-OS assists a customer with gift selection by analyzing dimensions and bag compatibility. It coordinates the Realman and Agilex robots, with the Agilex executing a VLA-cerebellum skill to ``open the gift bag,'' while the Realman selects and places the gift.
Future applications may explore more complex collaborations with three or more robots, significantly advancing embodied AI and robotics.

\section{Conclusion}
\label{sec:conclusion}
In this paper, we present \textbf{\textit{RoboOS}}, an open-source embodied system that improves multi-agent collaboration in industrial ecosystems. Utilizing a Brain-Cerebellum hierarchical architecture, \textbf{\textit{RoboOS}} overcomes challenges in adaptability and task scheduling. It features the Embodied Brain Model for decision-making, the Cerebellum Skill Library for skill execution, and Real-Time Shared Memory for coordination. This integration enables effective planning and error correction for complex tasks. Real-world experiments showcase RoboOS's versatility across various robotic embodiments, advancing embodied intelligence.
\clearpage
\textbf{Limitations}
\label{sec:limitations}
This paper focuses on experiments conducted in three specific environments: restaurants, households, and supermarkets. These settings were chosen to demonstrate the practical applications and effectiveness of our approach in everyday scenarios. However, due to limitations in our experimental environment, we were unable to explore additional contexts, such as factory settings and other industrial environments, including warehouses, assembly lines, and logistics hubs, where multi-robot collaboration is essential. These industrial environments often involve complex tasks that require coordinated efforts among multiple robots to achieve efficiency and productivity. Future work should aim to address these gaps and investigate the performance of our system in these critical scenarios, further showcasing its versatility and adaptability.




\bibliography{main}  

\clearpage
\appendix
\section*{Appendix}

This supplementary material provides additional details on the proposed method and experimental results that could not be included in the main manuscript due to page limitations.
Specifically, this appendix is organized as follows.

\begin{itemize}[left=1em]
\item Sec.~\ref{sec1} presents additional details of the models and training strategies. 
\item Sec.~\ref{sec2} presents details of training datasets and evaluation benchmarks.
\item Sec.~\ref{sec3} presents comprehensive experimental results on the performance of FlagScale~\cite{FlagScale2024}.
\end{itemize}

\section{Details of Models and Training}
\label{sec1}

RoboBrain-1.5-OS represents a significant advancement in robotic vision-language models, building upon the robust foundation of Qwen2.5-VL-7B~\cite{Qwen2.5-VL} through an elaborate three-stage training paradigm designed to progressively enhance both general and domain-specific capabilities.
(1) In Stage 1, the model undergoes full-parameter supervised fine-tuning (SFT) using 3M high-quality general VLM datasets with a learning rate of 1e-4, trained across 20 servers each equipped with 8$\times$A800 GPUs, aiming to establish robust foundational visual understanding and reasoning capabilities. (2) Stage 2 focuses on the robotics domain, utilizing 2.3M carefully curated robotic-related training data while retaining 10\% of Stage 1 data to prevent catastrophic forgetting. The learning rate is reduced to 1e-5 to ensure stable convergence. (3) The final Stage 3 specializes in optimization for RoboOS, first performing SFT with 245K OS-SFT samples (containing 2\% Stage 1 and 3\% Stage 2 data), followed by Group Relative Preference Optimization (GRPO)~\cite{shao2024grpo} using 4K OS-RL samples (only tool-calling samples), leveraging reinforcement learning (RL) for its efficiency and scalability. The GRPO phase in Stage 3 further reduces the learning rate to 1e-6, and completes training in 3 epochs on a single 8$\times$A800 server. The entire training process employs the DeepSpeed Zero3~\cite{zero} optimization strategy, with carefully configured key parameters including batch size (2 for SFT, 1 for GRPO), maximum sequence length (32768 for SFT, 8192 for GRPO), and weight decay (0.1). The number of completion (4) and maximum completion length (512) are specific for GRPO. Detailed hyperparameters are provided in Tab.~\ref{tab:training_setting}. This training scheme significantly enhances the model's performance in robotics applications and system compatibility while preserving the original capabilities of Qwen2.5-VL-7B~\cite{Qwen2.5-VL}.

\begin{table*}[!ht]
    \centering
    \caption{Detailed configuration for each training stage of the RoboBrain-1.5-OS.}
    \vspace{0.2cm}
    \label{tab:training_setting}
    \setlength{\tabcolsep}{12pt}
    \renewcommand{\arraystretch}{1.2}
    \resizebox{0.93\textwidth}{!}{%
    \begin{tabular}{@{}ll|c|c|c|c@{}}
        \toprule
        & & \textbf{Stage-1} & \textbf{Stage-2} & \multicolumn{2}{c}{\textbf{Stage-3}} \\ \cmidrule(l){3-3} \cmidrule(l){4-4} \cmidrule(l){5-6}
        & & \textbf{SFT} & \textbf{SFT} & \textbf{SFT} & \textbf{GRPO} \\
        \midrule 
        \multirow{2}{*}{\rotatebox[origin=c]{90}{\small \textit{Data}}}
        & \textbf{Dataset}  & General VLM & Robotic & OS-SFT (Part 1) & OS-RL (Part 2) \\
        & \#Samples & 3M & 2.3M & 245K & 4K \\
        \midrule 
        \multirow{2}{*}{\rotatebox[origin=c]{90}{\small \textit{Model}}}
        & \textbf{Trainable Part} & Full Model & Full Model & Full Model & Full Model \\
        & \#Tunable Parameters & 8.29B & 8.29B & 8.29B & 8.29B \\
        \midrule 
        \multirow{14}{*}{\rotatebox[origin=c]{90}{\small \textit{Training}}}
        & \textbf{Per-device Batch Size} & 2 & 2 & 4 & 1 \\
        & \textbf{Gradient Accumulation} & 2 & 2 & 2 & 2 \\   
        & \textbf{LR: $\{\psi_v^{\text{ViT}}, \phi_v^{\text{LLM}}\}$} & 1$\times 10^{-4}$ & 1 $\times 10^{-6}$ & 1 $\times 10^{-5}$ & 1 $\times 10^{-6}$ \\
        & \textbf{Epoch} & 1 & 1 & 1 & 3 \\
        & \textbf{Optimizer} & AdamW & AdamW & AdamW & AdamW \\
        & \textbf{Deepspeed} & Zero3 & Zero3 & Zero3 & Zero3 \\
        & \textbf{Weight Decay} & 0.1 & 0.1 & 0.1 & 0.0 \\
        & \textbf{Warmup Ratio} & 0.03 & 0.03 & 0.03 & 0.00 \\
        & \textbf{LR Schedule} & Cosine & Cosine & Cosine & Cosine \\
        & \textbf{Max Seq. Length} & 32768 & 32768 & 32768 & 8192 \\
        & \textbf{Max Compl. Length} & -- & -- & -- & 512 \\
        & \textbf{Num. of Compl.} & -- & -- & -- & 4 \\
        & \textbf{GPU Nums} & 20 $\times$ 8 & 20 $\times$ 8 & 4 $\times$ 8 & 1 $\times$ 8 \\
        \bottomrule
    \end{tabular}
    }
    \vspace{1mm}
    \vspace{-0.5em}
\end{table*}

\section{Details of Datasets and Benchmarks}
\label{sec2}

\subsection{Training Datasets}
The training of RoboBrain-1.5-OS leverages three comprehensive dataset categories: general VLM datasets for foundational capabilities, Robotic datasets for embodied intelligence, and RoboOS-Enhanced datasets for system-specific optimization.
The specific composition of these three datasets are listed as follows:

\begin{itemize}[left=1em]
    \item \textbf{General VLM Datasets} (Total: 3M samples) – We systematically organized five specialized subsets to establish fundamental capabilities:  
    (1) \textit{General-873k}, curated from LRV-400K~\cite{,liu2023mitigating} and LLaVA-665K~\cite{li2024llavaov} through rigorous filtering and restructuring, enhances broad question-answering with diverse QA pairs spanning descriptive, analytical, and inferential tasks.  
    (2) \textit{ScanView-318k} integrates multimodal 3D scene understanding data from MMScan-224K~\cite{mmscan} (annotated object segmentation and textual descriptions), 3RScan-43K~\cite{3rscan} (3D reconstructions with semantic labels), ScanQA-25K~\cite{scanqa} (QA pairs grounded in 3D environments), and SQA3D-26K~\cite{sqa3d} (spatial QA), enabling fine-grained environmental perception.  
    (3) \textit{VG-326k} combines Ref-L4~\cite{chen2024revisiting} (45K expressions spanning 365 object categories), OV-VG~\cite{li2024llavaov} (visual grounding smaples in Llava-OneVision), RefCOCO/RefCOCO+~\cite{yu2016modeling,mao2016generation} (natural language descriptions with restrictive/non-restrictive spatial constraints), and Visual Genome~\cite{krishna2017visual} (rich region descriptions and relational annotations) for precise visual grounding and localization.  
    (4) \textit{Spatial-R-1005k} leverages Ref-Reasoning-791K~\cite{yang2020graph} (complex expressions with attribute/spatial reasoning and adversarial images), COPS-Ref-148K~\cite{chen2020cops}, and Spatial-Trans-66K~\cite{tan2025reason} to model compositional relationships.  
    (5) \textit{Temporal-R-525k} synthesizes Embodied-Reasoner-9K~\cite{zhang2025embodied}, HandMeThat-300K~\cite{wan2022handmethat} (abstract command-based planning), and 200K+ simulated reasoning samples for sequential event comprehension. All datasets underwent deduplication and quality filtering to balance capability coverage while eliminating noise.

    \item \textbf{Robotic Datasets} (Total: 2.3M samples) – Designed to develop four core embodied intelligence competencies:  
    (1) \textit{Planning-700k} aggregates RoboVQA-Clean-200K (reconstructed from original RoboVQA-800K~\cite{sermanet2024robovqa}), ShareRobot-Plan-400K (a planning subset of ShareRobot~\cite{ji2025robobrain}), RoboBench-50K (constructed based on real-world embodied tasks by ourselves), and AgiBotWorld-Alpha-50K~\cite{bu2025agibot} to train hierarchical task decomposition and long-horizon planning.  
    (2) \textit{Pointing-537k} unifies Object-Ref-347K~\cite{yuan2024robopoint} (287K images with coordinate-based QA) and Pixmo-Point-190K~\cite{deitke2024molmo} (indoor scene point annotations) to refine spatial awareness via coordinate regression.  
    (3) \textit{Affordance-373k} merges Region-Ref-320K~\cite{yuan2024robopoint} (270K images with interactive region QA), PACO-LVIS-45K~\cite{ramanathan2023paco} (object functionality labels), and ShareRobot-Affordance-8K~\cite{ji2025robobrain} to predict actionable object properties.  
    (4) \textit{Trajectory-428k} combines LLaRVA-420K~\cite{niu2024llarva} and ShareRobot-Trajectory-8K~\cite{yuan2024robopoint} to learn manipulation sequences for successful execution. Each subset was iteratively optimized for robotic applicability, emphasizing precision in action-object mapping. To mitigate catastrophic forgetting while maintaining model performance, we implement a knowledge retention strategy where 10\% of Stage 1 training samples (Stage1-300K) are preserved during Stage 2.

    \item \textbf{RoboOS-Enhanced Datasets} (Total: 249k samples) – Tailored for RoboOS integration:  
    (1) \textit{Multi-Robot-45k} features 68 collaboration task types (supermarket/household/restaurant scenarios) generated via DeepSeek-V3-0324~\cite{liu2024deepseek}, with scene graphs, robot specs, and workflow visualizations for subtask reasoning.  
    (2) \textit{Robotic-Agent-144k} augments Multi-Robot-45k generated via DeepSeek-V3-0324~\cite{liu2024deepseek} with probabilistically sampled Observation-Action pairs (144K correct/error-injected variants) to improve operational robustness through SFT and RL training (140K for SFT and the rest for GRPO). This architecture ensures seamless adaptation to the RoboOS ecosystem implementation while preserving task-specific performance in multi-agent coordination and tool invocation.
\end{itemize}

\subsection{Evaluation Settings}
\textbf{Evaluation Baselines}
The reference baselines for comparison include: (1) Vision-Language Models (VLMs) of comparable parameter scales (e.g., LLaVA-OneVision-7B~\cite{li2024llavaov}, Qwen2.5-VL-7B~\cite{Qwen2.5-VL}, RoboBrain-7B-1.0~\cite{ji2025robobrain}, RoboPoint-14B~\cite{yuan2024robopoint}), and (2) larger general-purpose LLMs (e.g., Qwen3-14B~\cite{qwen3_2025}, DeepSeek-V3-685B~\cite{liu2024deepseek}). We evaluate vision-based and text-based benchmarks for VLMs, while only text-based benchmarks for LLMs. 

\begin{figure*}[!ht]
    \centering
   \vspace{-0.2cm}
    \includegraphics[width=0.93\linewidth]{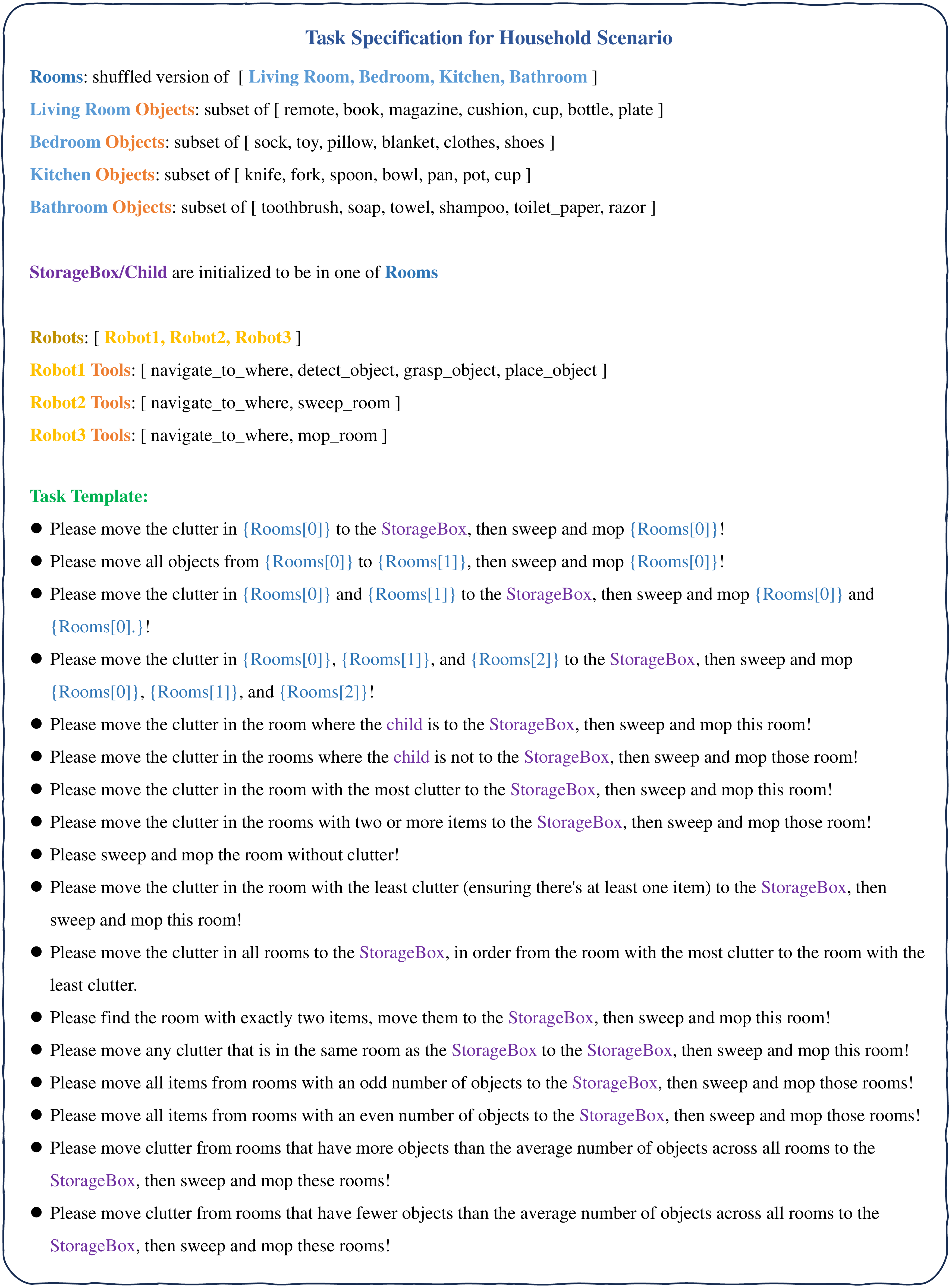}
    \vspace{-0.25cm}
    \caption{
        Task specification for household scenario. 
        }
       \vspace{-0.5cm}
    \label{fig:household}
\end{figure*}

\begin{figure*}[!ht]
    \centering
   \vspace{-0.2cm}
    \includegraphics[width=0.94\linewidth]{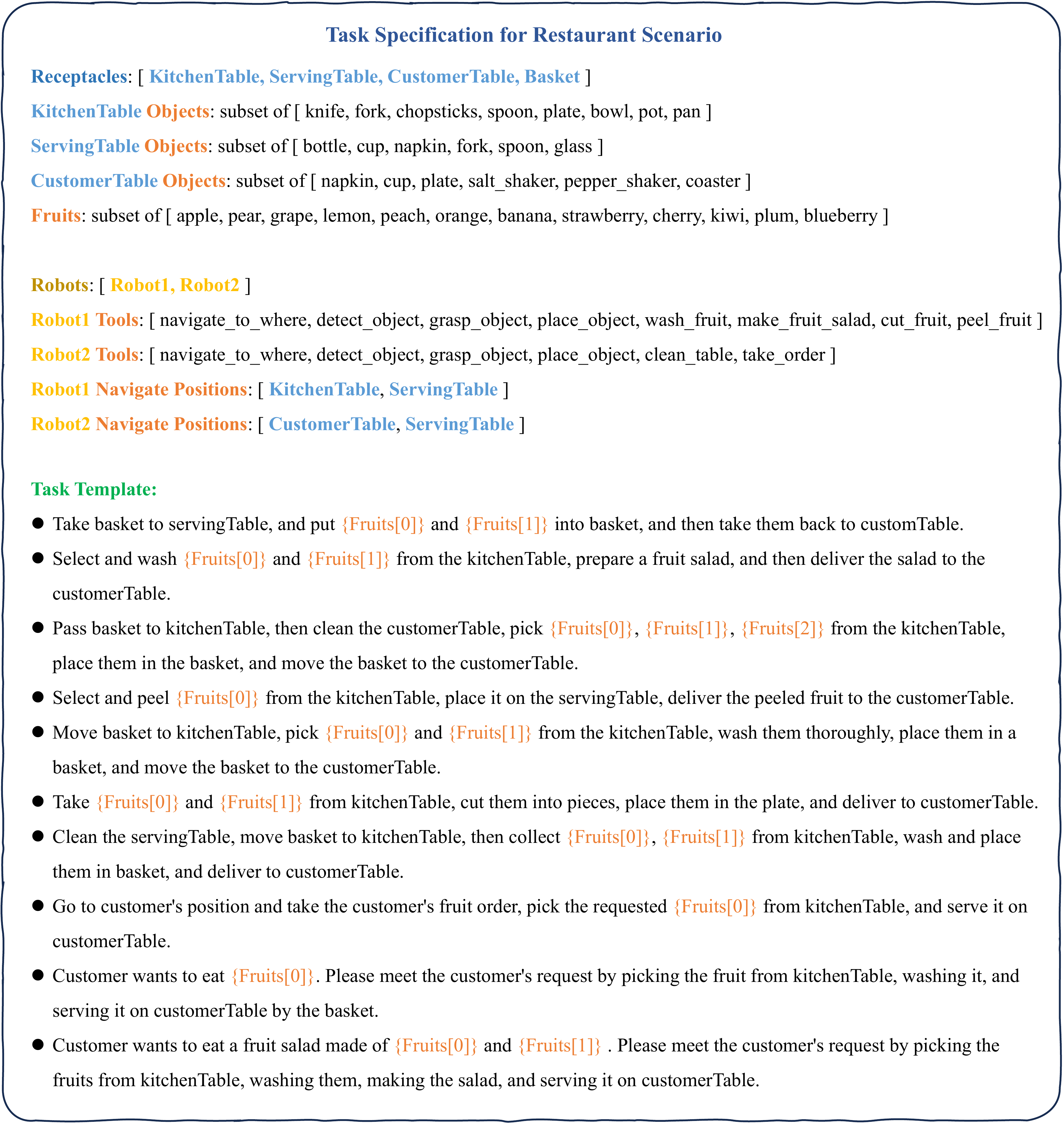}
    \vspace{-0.25cm}
    \caption{
        Task specification for restaurant scenario. 
        }
       \vspace{-0.5cm}
    \label{fig:restaurant}
\end{figure*}

\begin{figure*}[!ht]
    \centering
   \vspace{-0.2cm}
    \includegraphics[width=0.94\linewidth]{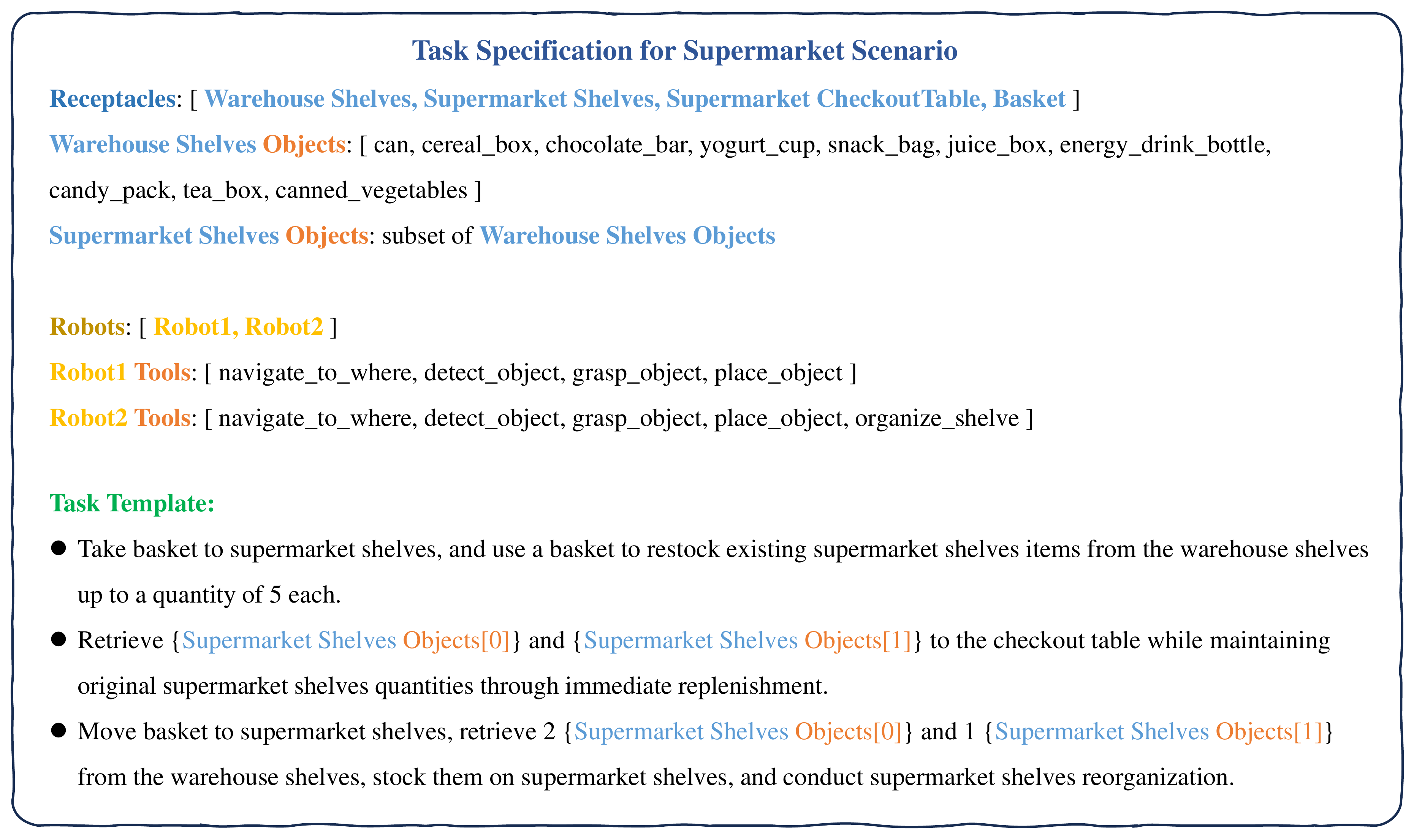}
    \vspace{-0.25cm}
    \caption{
        Task specification for supermarket scenario. 
        }
       \vspace{-0.5cm}
    \label{fig:supermarket}
\end{figure*}

\begin{figure*}[!ht]
    \centering
   \vspace{-0.2cm}
    \includegraphics[width=0.97\linewidth]{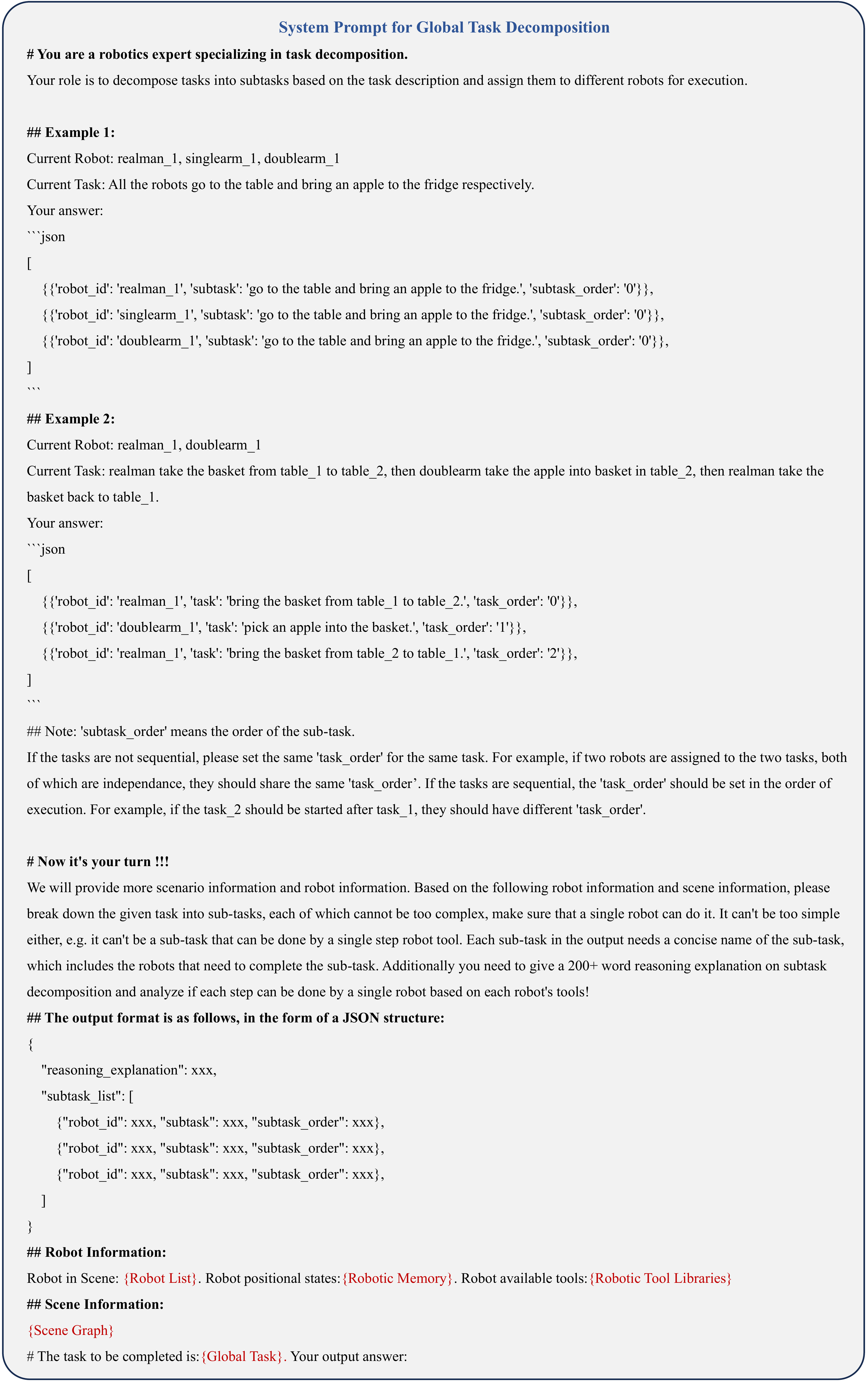}
    \caption{
        Prompt for global task decomposition. 
        }
       \vspace{-0.5cm}
    \label{fig:prompt1}
\end{figure*}

\begin{figure*}[!ht]
    \centering
   \vspace{-0.2cm}
    \includegraphics[width=0.97\linewidth]{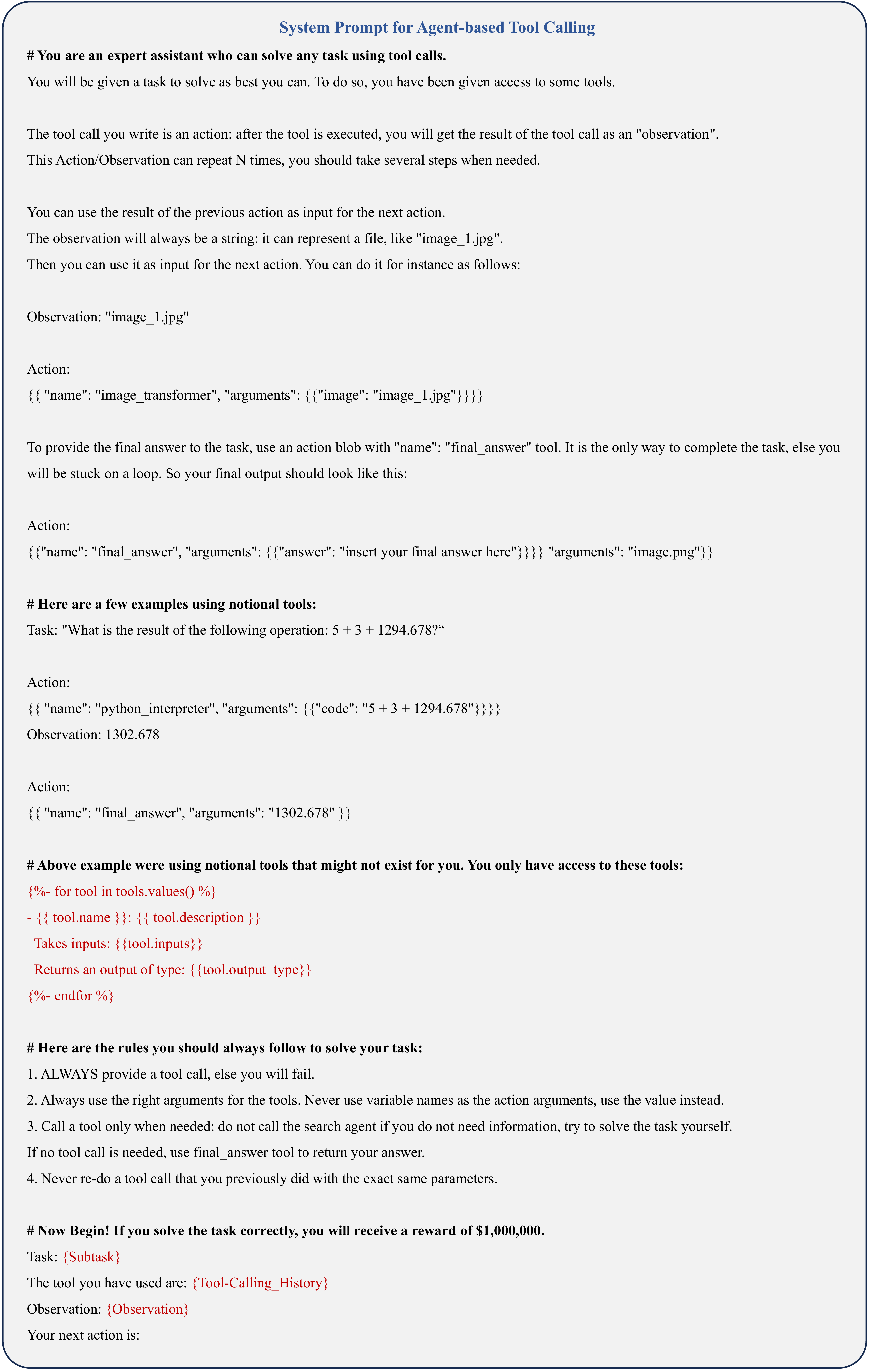}
    \caption{
        Prompt for agent-based tool calling. 
        }
       \vspace{-0.5cm}
    \label{fig:prompt2}
\end{figure*}

\textbf{Evaluation Benchmarks}
We conducted comprehensive evaluations on the model's four core robotic operation capabilities (Multi-Robot Planning, Pointing Prediction, Affordance Prediction, and Trajectory Prediction), with the benchmark configurations specified as follows:

\begin{itemize}[left=1em]
\item \textbf{Multi-Robot Planning} 
Our evaluation framework assesses multi-robot planning capabilities across three task scenarios: restaurant environments, commercial supermarkets, and household settings. Using RoboOS as the testbed system, we employ the Tool-Calling Accuracy Rate (AR) metric~\cite{zhang2024lmms} for quantitative assessment. For the three task scenarios of restaurant, supermarket, and household settings, we randomly sampled and generated 200 task samples for each scenario as a benchmark for testing. The task specifications for each scenario are shown in Fig.~\ref{fig:household}-\ref{fig:supermarket}. These samples are used to evaluate global task decomposition and agent-based tool-calling in RoboOS, with corresponding prompts illustrated in Fig.~\ref{fig:prompt1} and Fig.~\ref{fig:prompt2}.

\item \textbf{Pointing Prediction} 
We evaluate pointing prediction performance using the Where2Place dataset~\cite{yuan2024robopoint}, which contains 100 real-world images depicting cluttered environments with annotated spatial relations. Each image includes: (i) a textual description specifying desired free space, (ii) a ground-truth mask of the target region, and (iii) corresponding query points that probe the model's ability to localize referenced spaces. Performance is quantified through Hit Accuracy measured against target masks using the AR metric, assessing the precision of robotic systems in interpreting spatial references and predicting intended pointing locations.

\item \textbf{Affordance Prediction}
We utilize the AGD20K benchmark~\cite{luo2024grounded} - a comprehensive dataset comprising over 20,000 annotated images spanning diverse affordance categories - to evaluate the performance of affordance prediction. The evaluation protocol measures mean Average Precision (mAP) across multiple Intersection-over-Union (IoU) thresholds (0.25, 0.50, 0.75, 0.90), providing rigorous assessment of model robustness.

\item \textbf{Trajectory Prediction}
Our trajectory analysis employs the ShareRobot-Trajectory benchmark~\cite{ji2025robobrain} with three complementary metrics: 
(1) Discrete Fréchet Distance (DFD)~\cite{gu2023rt}: Computes the minimum leash length required for coupled traversal of predicted and ground-truth trajectories, capturing both geometric similarity and temporal alignment.
(2) Hausdorff Distance (HD): Measures worst-case positional deviation between trajectories.
(3) Root Mean Square Error (RMSE): Quantifies average pointwise Euclidean distance errors.
This multi-metric approach enables hierarchical analysis of trajectory quality, from global shape preservation (DFD) to local precision (RMSE), with HD identifying critical failure cases.

\end{itemize}

\section{Efficiency Performance on FlagScale}
\label{sec3}
This study conducts a comprehensive evaluation of the impact of FlagScale~\cite{FlagScale2024} on inference efficiency within the RoboOS framework through controlled comparative testing between two system configurations: RoboOS without FlagScale \textcolor{orange}{(Baseline)} and RoboOS with FlagScale  \textcolor{blue}{(+ FlagScale)}.

\subsection{Experimental Setup}
 All controlled comparative testing experiments were performed on an NVIDIA RTX 4090 GPU featuring 24GB of GDDR6X VRAM, noting that this hardware architecture lacks native support for FP8 tensor operations. The evaluation environment maintained strict parameter consistency, including fixed GPU memory utilization at 90\% through the gpu\_memory\_utilization=0.9 setting while deliberately disabling prefix caching to establish baseline performance measurements.
The experimental design incorporated several optimized computational parameters to ensure representative benchmarking conditions. These included a maximum sequence limit of 16 concurrent processes (max\_num\_seqs=16) and a progressive CUDA graph capture strategy configured for batch sizes [1, 2, 4, 8, 16], enabling efficient batch processing across varying workload demands. Primary performance metrics focused on End-to-Token Latency (E2TL) measurements for 85-token generation outputs, with particular attention to the characteristic disparity between initial batch latency and steady-state operational performance. The evaluation further examined scaling characteristics through Tensor Parallelism configurations at three distinct levels (TP1, TP2, and TP4) to assess both single-device and multi-device acceleration potential.

\subsection{FP16 Performance}
The FP16 latency comparisons demonstrate consistent acceleration benefits from FlagScale optimization across all tensor parallelism configurations. As shown in Tab.~\ref{tab:latency_comparison1}, the relative improvement decreases monotonically from 48.3\%–62.9\% at batch size of 1, to 16.4\%–20.8\% at batch size of 16 under TP1–TP4 configurations. This inverse correlation between batch size and optimization efficacy suggests diminishing returns for memory-bound operations at larger workloads. The non-linear latency growth pattern—where baseline latency increases 3.18× from batch 1 to 16 versus 5.14× for FlagScale—indicates superior batch processing scalability of the optimized implementation.

\begin{table}[h]
\centering
\caption{FP16 Latency Comparison (85-token generation, ms)}
\label{tab:latency_comparison1}
\resizebox{0.98\columnwidth}{!}{
\begin{tabular}{c|cc|cc|cc}
\toprule
\multirow{2}{*}{Batch Size} & \multicolumn{2}{c|}{E2TL-TP1} & \multicolumn{2}{c|}{TP2} & \multicolumn{2}{c}{TP4} \\
\cmidrule(lr){2-3} \cmidrule(lr){4-5} \cmidrule(lr){6-7}
 & \textcolor{orange}{Baseline} & \textcolor{blue}{+ FlagScale} & \textcolor{orange}{Baseline} & \textcolor{blue}{+ FlagScale} & \textcolor{orange}{Baseline} & \textcolor{blue}{+ FlagScale} \\
\midrule
1 & 4.06 & 2.10 (-48.3\%) & 3.85 & 1.52 (-60.5\%) & 3.29 & 1.22 (-62.9\%) \\
2 & 4.64 & 2.69 (-42.0\%) & 4.41 & 2.08 (-52.8\%) & 3.96 & 1.75 (-55.8\%) \\
4 & 5.79 & 3.83 (-33.9\%) & 5.24 & 3.16 (-39.7\%) & 4.95 & 2.78 (-43.8\%) \\
8 & 8.09 & 6.13 (-24.2\%) & 7.34 & 5.25 (-28.5\%) & 7.31 & 4.90 (-33.0\%) \\
16 & 12.92 & 10.80 (-16.4\%) & 12.16 & 9.63 (-20.8\%) & 11.21 & 9.11 (-18.7\%) \\
\bottomrule
\end{tabular}
}
\end{table}

\subsection{W8A16 Performance}
Quantization to W8A16 precision yields additional latency reductions beyond FP16 baselines, with FlagScale achieving 54.0\%–65.2\% improvement for single-query inference, as shown in Tab.~\ref{tab:latency_comparison2}. The TP4 configuration maintains the most significant gains across all batch sizes, demonstrating 23.4\%–65.2\% lower latency compared to baseline. Two key observations emerge: (1) The absolute latency values under W8A16 are consistently 12\%–17\% lower than corresponding FP16 results, confirming the expected quantization benefits; (2) The optimization maintains stable relative improvements across precision modes, with TP2 showing particularly consistent 20.0\%–60.5\% reductions. The convergence of latency values at batch size 16 (9.11–10.62ms across TP modes) suggests a hardware-bound limit for large-batch processing regardless of parallelism configuration.

\begin{table}[h]
\centering
\caption{W8A16 Latency Comparison (85-token generation, ms)}
\label{tab:latency_comparison2}
\resizebox{0.98\columnwidth}{!}{
\begin{tabular}{c|cc|cc|cc}
\toprule
\multirow{2}{*}{Batch Size} & \multicolumn{2}{c}{E2TL-TP1} & \multicolumn{2}{c}{TP2} & \multicolumn{2}{c}{TP4} \\
\cmidrule(lr){2-3} \cmidrule(lr){4-5} \cmidrule(lr){6-7}
& \textcolor{orange}{Baseline} & \textcolor{blue}{+ FlagScale} & \textcolor{orange}{Baseline} & \textcolor{blue}{+ FlagScale} & \textcolor{orange}{Baseline} & \textcolor{blue}{+ FlagScale} \\
\midrule
1  & 3.48 & 1.60 (-54.0\,\%) & 3.34 & 1.32 (-60.5\,\%) & 3.36 & 1.17 (-65.2\,\%) \\
2  & 4.17 & 2.22 (-46.8\,\%) & 4.21 & 1.91 (-54.6\,\%) & 3.92 & 1.69 (-56.9\,\%) \\
4  & 5.52 & 3.39 (-38.6\,\%) & 5.08 & 2.99 (-41.1\,\%) & 5.28 & 2.76 (-47.7\,\%) \\
8  & 7.78 & 5.78 (-25.7\,\%) & 7.46 & 5.25 (-29.6\,\%) & 6.96 & 4.84 (-30.5\,\%) \\
16 & 12.98 & 10.62 (-18.2\,\%) & 12.53 & 10.03 (-20.0\,\%) & 11.90 & 9.11 (-23.4\,\%) \\
\bottomrule
\end{tabular}
}
\end{table}

\end{document}